\documentclass[runningheads,a4paper]{llncs}

\usepackage{array}
\usepackage{multirow}
\usepackage{color}
\usepackage{amssymb}
\usepackage{amsmath}
\usepackage{dsfont}
\usepackage{mathrsfs}
\usepackage{graphicx}
\usepackage{url}
\usepackage{bm}
\usepackage{subfigure}
\usepackage{siunitx}
\usepackage{graphicx}
\graphicspath{{./pdf/}{./jpg/}{./png/}}
\DeclareGraphicsExtensions{.pdf,.jpg,.png}

\begin{document}
\mainmatter

\title{Large-scale mammography CAD with Deformable Conv-Nets}
\titlerunning{Large-scale mammography CAD with Deformable Conv-Nets}
\toctitle{Large-scale mammography CAD with Deformable Conv-Nets}

\author{Stephen Morrell\inst{1}
\email{stephen.morrell@gmail.com}
\and Zbigniew Wojna\inst{1} 
\and Can Son Khoo\inst{1} 
\and Sebastien Ourselin\inst{2} 
\and Juan Eugenio Iglesias\inst{1} }

\authorrunning{Morrell, Wojna, Khoo, Ourselin and Iglesias}
\institute{Medical Physics and Biomedical Engineering, University College London, UK \\
\and Sch. Biomed. Eng. Im. Sci., King's College London}
\tocauthor{Morrell, Wojna, Khoo, Ourselin and Iglesias}

\maketitle

\begin{abstract}
State-of-the-art deep learning methods for image processing are evolving into increasingly complex meta-architectures with a growing number of modules. Among them, region-based fully convolutional networks (R-FCN) and deformable convolutional nets (DCN) can improve CAD for mammography: R-FCN optimizes for speed and low consumption of memory, which is crucial for processing the high resolutions of to \SI{50}{\micro\metre} used by radiologists. Deformable convolution and pooling can model a wide range of mammographic findings of different morphology and scales, thanks to their versatility. In this study, we present a neural net architecture based on R-FCN / DCN, that we have adapted from the natural image domain to suit mammograms -- particularly their larger image size -- without compromising resolution. We trained the network on a large, recently released dataset (Optimam) including 6,500 cancerous mammograms. By combining our modern architecture with such a rich dataset, we achieved an area under the ROC curve of 0.879 for breast-wise detection in the DREAMS challenge (130,000 withheld images), which surpassed all other submissions in the competitive phase. 
\end{abstract}

\section{Introduction}
\label{sec:intro}
Breast cancer is the most commonly diagnosed cancer and the second leading cause of cancer death in U.S. women~\cite{acs}. Timely and accurate diagnosis is of paramount importance since prognosis is improved by early detection and treatment, notably before metastasis has occurred. Screening asymptomatic women with mammography reduces disease specific mortality by between 20\% and 40\%~\cite{acs} but incorrect diagnosis remains problematic. Radiologists achieve an area under the ROC curve (AUC) between 0.84 and 0.88~\cite{Lehman2015}, depending on expertise and use of computer aided detection (CAD).

CAD for mammography was first approved 20 years ago but some studies showed it to be ineffective~\cite{Gross} or counterproductive~\cite{Lehman2015} because of over-reliance. Early CAD methods used simple handcrafted features and produced many false positive detections~\cite{Lehman2015}. The best of these ``classical'', feature-engineered methods, represented by e.g.,~\cite{Jiang2016,Karssemeijer1996}, plateaued at 90\% sensitivity for masses at one false positive per image~\cite{Karssemeijer1996}, and at 84\% area of overlap in segmentation~\cite{Jiang2016}. 

Deep learning (DL) has enhanced image recognition tasks, building on GPUs, larger data sets and new algorithms. Convolutional neural nets (CNNs) have been applied to mammography, outperforming classical methods. For example, Dhungel et al.~\cite{Dhungel16} used CNNs to achieve state of the art results in mass classification. In a recent study, Kooi et al.~\cite{Kooi2017} proposed a two-stage system in which a random forest classifier first generated proposals for suspicious image patches, and a CNN then classified such patches into malignant or normal groups. Kooi's system was trained on a large private dataset of 40,506 images (6,729 cases) of which 634 were cancerous, and achieved an AUC of 0.941 in patch classification -- representing significant improvement beyond prior work.

Despite the adoption of CNNs and increasing size of datasets, mammographic analysis still lags work on natural images in dataset size and algorithm comparability. Databases like ImageNet~\cite{ILSVRC15}  and MS COCO~\cite{lin2014microsoft} include millions of instances. Moreover, these public datasets enable independent verification of algorithmic performance with private test sets. Very recently, the Optimam~\cite{optimamonline} and Group Health datasets  have begun to approach ImageNet and MS Coco sizes. Group Health was made available under the DREAMS Digital Mammography Challenge~\cite{dream} (henceforth ``the Challenge''), which used a verified hidden test set to benchmark comparisons between methods. The Challenge had 1,300 participants, and was supported by the FDA and IBM among others.  

Moreover, CNN architectures now include a plethora of new techniques for detection, classification and segmentation~\cite{dai2017deformable,lin2017feature,he2017mask}. It has also been shown that integration of these tasks in unified architectures -- rather than pipelining networks --  not only enables more efficient end-to-end training, but also achieves higher performance than when tasks are performed independently (e.g.,~\cite{he2017mask}). This is due to sharing of features that use richer locality information in the labels -- segmentations or bounding boxes.

Here we present our submission to the second phase (``collaborative'') of the Challenge. Our contribution includes the selection of architectures from the natural image domain, adaptation to mammography to balance the trade-off between high resolution and network size (computational tractability) for fine feature detection, e.g., microcalcifications, data augmentation and score aggregation. In particular, the presented system is -- to the best of our knowledge -- the highest resolution DL mammography object detection system ever trained.

\section{Methods}
\label{sec:methods}

\subsection{Network architecture}
Even though our objective is classifying whole images, we chose a detection architecture to exploit the rich bounding box information in our training dataset (Optimam), and also to increase the interpreteability of the results, which is useful for clinicians. Our choice of meta-architecture is Region-based Fully Convolutional Networks (R-FCN)~\cite{Networks}, which are more memory-efficient than the popular Faster Region-based Convolutional Neural Nets (F-RCNN)~\cite{ren2015faster}. R-FCNs were enhanced with Deformable Convolutional Networks (DCN)~\cite{dai2017deformable}, which  dynamically model spatial transformations for convolutions and Regions of Interest (ROI) Pooling, depending on the data's current features: 
$$
y(\bm{p}_0) = \sum_{\bm{p}_n \in \mathcal{R}} \bm{w} (\bm{p}_n) \cdot \bm{x}(\bm{p}_0+\bm{p}_n  + \Delta \bm{p}_n),
$$
where $y$ is the filter response at a location $\bm{p}_0$; $\mathcal{R}$ is a neighborhood around $\bm{p}_0$; and $\bm{w}$ and $\Delta \bm{p}_n$ are learnable sets of weights and offsets, respectively. The versatility of adaptive convolution and pooling enables DCNs to model a wider spectrum of shapes and scales, which is appropriate in mammography -- where features of interest can be of very different sizes (from barely perceptible microcalcifications to large masses) and forms (foci, asymmetries, architectural distortions). 

\begin{figure}[!t]
\centering
\includegraphics[width=1.0\textwidth]{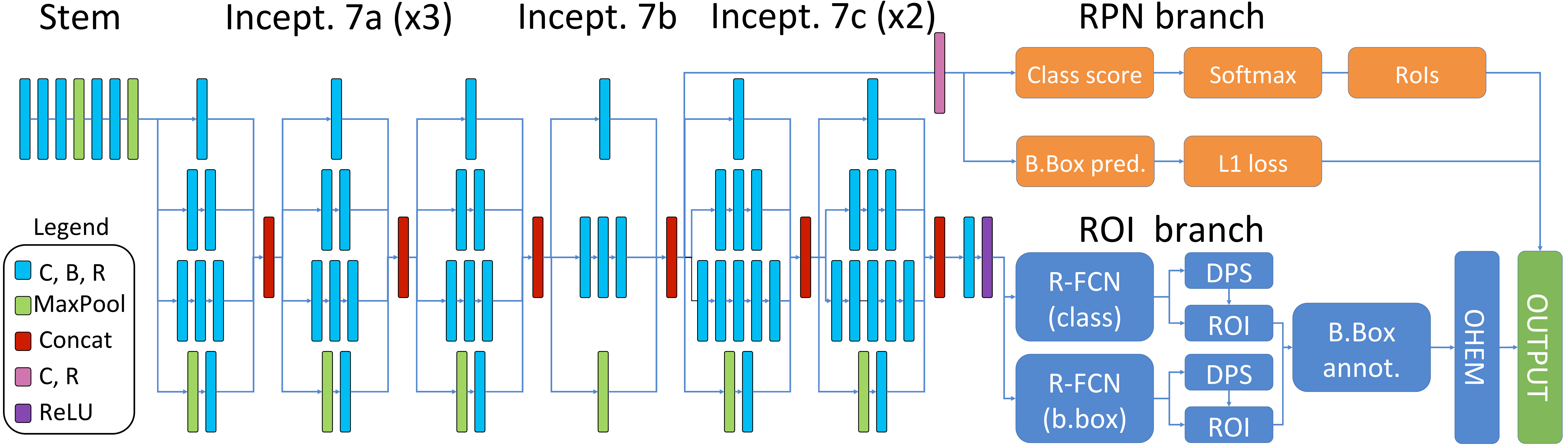}
\caption{Network architecture used in this study. C, B and R stand for convolution, batch norm and ReLU layers; RPN for region proposal network;  DPS (ROI) for deformable position sensitive ROI; and OHEM~\cite{shrivastava2016training} for online hard example mining. }
\label{fig:architecture}
\end{figure}

A diagram of our architecture is shown in Fig.\ref{fig:architecture}. It starts with a detection backbone, followed by two parallel branches: a region proposal network (RPN) branch, and a region of interest (ROI) branch -- as per the R-FCN meta-architecture. The RPN branch~\cite{ren2015faster} proposes candidate ROIs, which are applied on the score maps from the Inception 7b module. The ROI branch uses deformable position sensitive (DPS) score maps to generate class probabilities. Analogously to deformable convolutions,  deformable ROI pooling modules include a similar parallel branch (Fig. 4 in~\cite{dai2017deformable}), in order to compute the offsets. Deformable pooling can directly replace its plain equivalent and can be trained with back propagation.

\subsection{Adaptations}
\label{sec:adapt}
\noindent\textbf{Backbone:}  
Our backbone is descended from Inception v3~\cite{Szegedy15} from which we selected the first 7 layers (the ``stem'') and  modules 7A, 7B and 7C. Choosing Inception, for which pre-trained weights from natural images were published, allowed transfer learning which showed beneficial for mammographic image analysis~\cite{carneiro}. We included early layers on the assumption these are more consistent between domains. We chose consecutive layers to preserve co-adaption of weighs where possible. We compared to other recent architectures~\cite{Hu2017a,Huang2016a,Xie2016} in a pilot dataset but results were weaker; we did not pursue them.

\noindent\textbf{Resolution-related trade-offs:}  
Current GPU memory constraints preclude full size mammographic images in deep CNNs -- yet radiologists regularly zoom in to the highest level of detail. In particular, malignant microcalcifications may only be discerned at $\sim$\SI{50}{\micro\metre} resolution (approximately $4,000 \times 5,000$ pixels). Leading CNNs are designed for maximal GPU memory usage when fed natural images, which have two orders of magnitude fewer pixels, so the CNNs must be trimmed judiciously for mammography.  This results in a trade-off between backbone choice, module selection, image downsampling, batch size, and number of channels in the different layers. We selected Inception v3 for its superior trade-off between parameter parsimoniousness and accuracy in natural images~\cite{Canziani2016}. Its successor, Inception ResNet v2, was used in~\cite{dai2017deformable} but would have restricted images to $1,300 \times 1,300$ pixels. We included fewer repeats of modules 7A, 7B and 7C: three, one and two repeats, respectively; pilot experiments showed fewer layers to be sufficient for mammograms, whose content is less heterogeneous than natural images. We reduced batch size to one image per GPU. The channels are co-adapted in the pretrained weights and we ultimately retained them all. These choices, combined with meta-architecture and framework choices, enabled input size of $2,545 \times 2,545$ pixels (i.e. minimum downsample factor of 0.42), the highest resolution used for mammography classification to the best of our knowledge. 
% Experiments with different input image sizes are included in Section ~\ref{sec:expAndRes}.

\noindent\textbf{Data augmentation: } 
Training is more effective if additional augmented data are included. Each training image was rotated through 360$^{\circ}$ in 90$^{\circ}$ increments and flipped horizontally, thus included eight times per epoch. We opted for the benefit of using rotated images despite induced ``anatomical'' noise, i.e., implausible anatomy. We also used the same four rotations and flipping at inference. We did not use random noise or random crops (which are standard in the natural image domain), to avoid omitting lesions at the image edge. % (a common site for radiologist misses). 

\noindent\textbf{Aggregation of multiple views:} 
Screening exams usually consist of two views (cranio-caudal, CC, and medio-lateral oblique, MLO) of each breast, giving four images per exam. For each of these images we generated 8 predictions, when including augmentations. Each subjects's probability of malignancy  lesion was calculated by computing  the mean over views and augmentations for each laterality, and then taking the maximum over the two sides; other combination rules were explored but yielded inferior results (see results in Section~\ref{sec:results}). 

% \noindent\textbf{Other adaptations:} We adjusted the Non Maximum Suppression and Intersection over Union thresholds and other parameters.

\section{Experiments and results}
\label{sec:expAndRes}

\subsection{Data} 
Two datasets were released in 2016/17, which were substantially larger than prior digital mammography datasets.  % yet together have only one citation of which we are aware. 
These are the Optimam~\cite{optimamonline} dataset, which we used in training, and Group Health (GH), used in testing. The ground truths were determined by biopsy. Using standardised data makes comparisons objective and reproducible, e.g., as for ImageNet in the natural image domain. We believe these new benchmarks will allow attribution to future architectures. 

Group Health images are a representative sample of 640,000 screening mammogram images, approximately 0.5\% cancerous, provided in the Challenge. All machines were Hologic and maximum image size was $3,300 \times 4,100$ pixels. The GH data were not downloadable, excepting a small pilot set of 500 images for prototyping. The data are kept on IBM's cloud and are not accessible directly. Challenge participants could only upload models, run inference on the cloud, and receive a score. While this hampers testing experiments, it preserves patient confidentiality and ensures veracity of results. We used two subsets of GH in our study: 1. GH-13K, a subset of approximately 13,000 images with cancer prevalence inflated artificially by a factor of four; and 2. GH-Validation, a representative subset with 130,000 images, which was used for final testing and ranking and which the organisers intend to keep open for future testing, providing a hitherto absent way to benchmark performance. 

Optimam consists of 78,000 selected digital screening and symptomatic mammograms including approximately 7,500 findings with bounding boxes of which 6,500 were cancerous. It included preprocessed and magnification images. Mammography machines were mainly Hologic and GE, and maximum image size was $4,000 \times 5,000$ pixels. Most teams in phase 2 of the Challenge used Optimam. % collaborators -> participants   saves a line! :-)

\subsection{Experimental setup}

We trained our network on all Optimam images with findings. We trained our architecture on three different classes: negative, benign and malignant findings. For subsequent analyses, we used the score of the malignant class as prediction score. Pixel intensities were normalised across different manufacturers and devices using the corresponding lookup tables. We chose MXNet~\url{http://mxnet.incubator.apache.org/}, for its memory-efficiency which surpasses most frameworks including TensorFlow. In terms of parameters, we used the same values as in the original publications describing the different DL modules, with two main differences: 1. We changed the scale of the input images, as explained in Section~\ref{sec:adapt}; and 2. In order to reflect the much lower risk of overlap and occlusion observed in mammography compared with natural images, we reduced the RPN positive overlap parameter from 0.7 to 0.5, and the proposal NMS threshold from 0.7 to 0.1. Training took 48h on two NVIDIA TitanX GPUs. 

We chose AUC on breast or patient classification as measure of diagnostic accuracy. AUC is used frequently to estimate the diagnostic performance of both CAD and radiologists. Compared with metrics like specificity at sensitivity or partial AUC, which are usually applied at high sensitivity levels and may be used to evaluate screening radiologists,  AUC measures diagnostic accuracy across all probability thresholds, so is a comprehensive metric.
% and it is used more often than the precision / recall area under the curve. 
Use cases for DL algorithms may range form automated flagging of some positive cases for immediate follow up -- where high precision at high confidence thresholds is key -- to safely excluding normal mammograms -- where high negative predictive value at low confidence thresholds prevails.

We conducted three sets of experiments. First, we tested a number of design choices on the pilot set, then second on the GH-13K dataset, changing one or two key variables at a time while holding others constant. We tested backbone choices, scales, train and test augmentation and aggregation. Each evaluation on GH-13K took a day on the IBM cloud. Finally, we submitted our final model for testing on GH-Validation, which took approximately 8 days.

\subsection{Results}
\label{sec:results}

Table~\ref{table:accuracy} summarises results from our GH-13K experiments. Experiments 1-5 explored combinations of backbones and scales. In terms of backbone, empirical results showed that Inception was superior to ResNet~\cite{he2016deep} (1 vs. 2 and 5). In terms of scales, the main conclusion was that performance peaked when train and test inference was run at the higher scale (2, 8); however, accuracy dropped when testing size exceeded training (3, 4). Experiments 6-8 assessed the impact of augmentation (6), as well as answering the question of how to aggregate augmentation scores (7). In general, these experiments showed that: a) augmentation at inference does help; and b) within a single breast, taking the mean over views and augmented images outperformed the alternative of taking the maximum (we still use the maximum across lateralities).  
% Many comparisons are to our final model with AUC 0.8667 (variance 8.35e-5) and are included despite high \textit{p}-values. 
Row 9 shows a test AUC of 0.74 on GH-Validation from the competitive phase of the Challenge using a classification net (an adapted Inception ResNet v2), TensorFlow with the largest possible resolution under that setup (1,600$\times$1,600, much smaller than our current model, thus leading to a 0.15 decrease in AUC). 
% We omit the many pilot experiments due to lack of space. 

% question for Eugenio: should the caption be above the table?
\begin{table}[t!]
\caption{Summary of GH-13K results. Legend for consistent variables: R - ResNet backbone; M - mean probability over augmentations and views, maximum over laterality; I - Inception backbone; N - No augmentation at inference. The scale is in pixels.}
\begin{tabular}{| c | c | c | c | c | c |}
\hline
	Exp. \# & Changed variable  & AUC & AUC  & Consistent \\
	   &   &  before&  after &  variables\\
	\hline
1	&	Train and Test Scale: 2145 to 2545	&	 0.8352 	&	 0.8227	&	R, M, N	\\
2	&	Train and Test Scale: 2145 to 2545	&	 0.8595 	&	 0.8667 	&	M, I	\\
3	&	Increased test image size: 2500 to 2900	&	 0.8173 	&	 0.8143	&	I	\\
4	&	Increased test image size: 2900 to 3300	&	 0.8143 	&	 0.8039	&	I	\\
5	&	Changed backbone: Resnet to Inception	&	 0.8352 	&	 0.8584	&	M, N\\
6	&	Added augmentation at inference	&	 0.8584 	&	 0.8667 &	M, I  \\
7	&	Max over breast's images changed to mean &	 0.8591 	&	 0.8667 & I\\
8	&	As for 8 above and inference scale: 2,454 to 2,545	&	 0.8511 	&	 0.8667 	&	I	\\
9	&	Adapted Inception ResNet v2. Image size of $1,600\times1,600$& N/A & 0.7366 & M, I, N \\
\hline 
\end{tabular}
\label{table:accuracy}
\end{table}

\begin{figure}%[!t]
\centering
\subfigure[]{
\includegraphics[width=.175\textwidth]{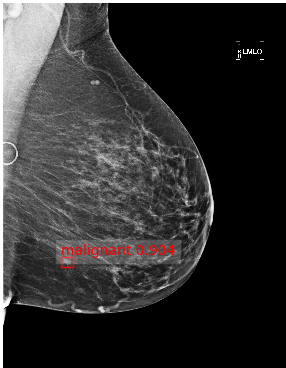} 
}
\subfigure[]{
\includegraphics[width=.175\textwidth]{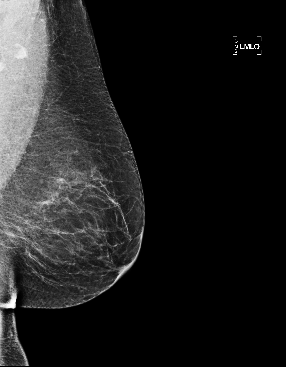}
}
\subfigure[]{
\includegraphics[width=.175\textwidth]{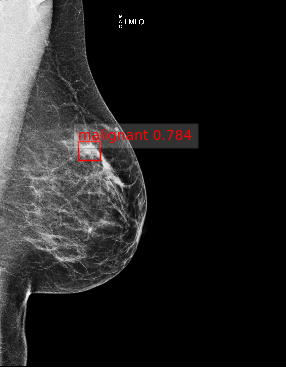}
}
\subfigure[]{
\includegraphics[width=.175\textwidth]{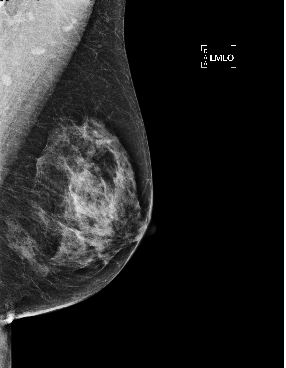}
}
\subfigure[]{
\includegraphics[width=.175\textwidth]{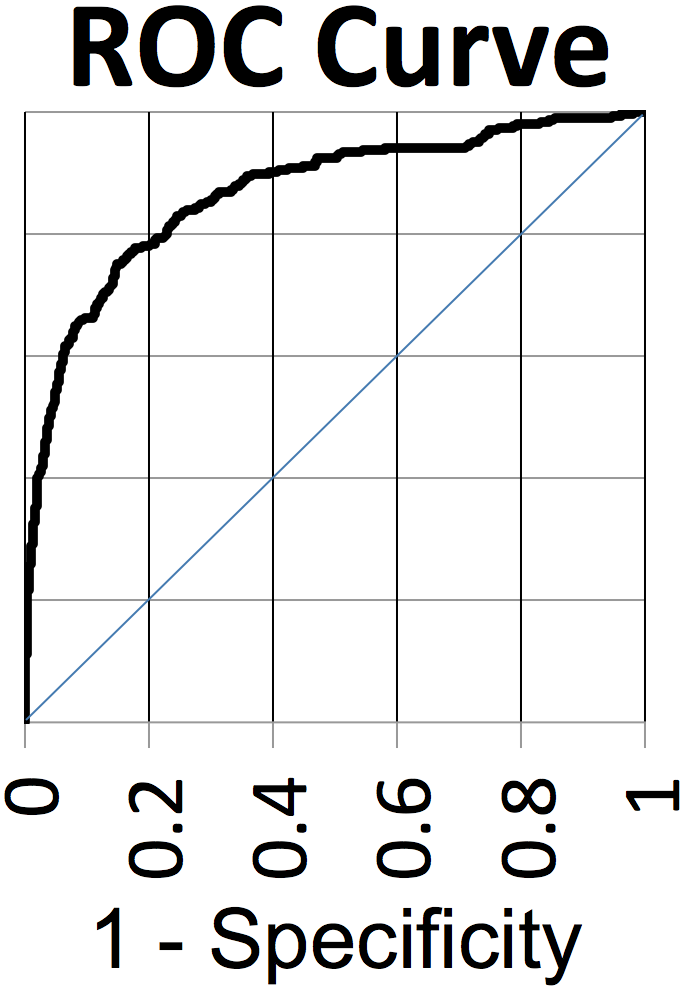}
}
\caption{(a)~True positive prediction ($p=0.90$ probability of malignancy) of an inconspicuous lesion on a left MLO of a 73 year old woman. (b)~True negative (malignancy: $p=0.06$) for left MLO view of a 66 year old woman. (c)~False positive ($p=0.78$) on left MLO of a 43 year old woman, due to hyper-intense region. (d)~False negative ($p=0.03$) for left CC view of a 61 year old woman. (e) ROC by breast, AUC = 0.879.}
\label{fig:samplesSynthetic}
\end{figure}

Based on these results, we submitted our final architecture described in Section~\ref{sec:methods} for testing on the large GH-Validation dataset. In the first sub-challenge, which records the AUC by breast purely on imaging and blinded to demographic information, we achieved AUC=0.879 (standard deviation: 0.00914), see Fig. 2(e), which is 0.005 above the top AUC in the competitive phase of the Challenge. It was also the highest single-model AUC in the collaborative phase, 0.014 below an ensemble of detection models, and higher than all patch-based models. The second sub-challenge is on subject-wise AUC, with access to both images and demographics. Despite ignoring demographics, our architecture gave AUC=0.868, behind only the top score in the competitive phase (a patch-based curriculum-trained model) by 0.006. Twenty-five method descriptions from this phase are available at \url{synapse.org}, but details of the collaborative phase, including performance of patch-based models trained on Optimam, is embargoed pending publication by the Challenge. Fig.~\ref{fig:samplesSynthetic} shows sample outputs from GH. %, including true and false positive and negatives.

%width=.226 ok for 4. .181 gives 4 .170 gives 5

\section{Discussion and conclusion}
\label{sec:conclusion}

We have presented a two-stage detection network trained on strongly labelled data which achieved 0.879 AUC by breast on a large unseen representative screening test set, operating at high resolution. Important questions remain, e.g., the impact of deformable modules, ameliorating batch size = 1 in batch normalisation, image rotation, the use of architectures that handle multiple scales (FPNs), single pass models, comparison to the different setup in~\cite{Kooi2017} and clinical applicability. Exploring these directions, along with integrating demographic features into the architecture, will be in a future journal extension. As mammographic training databases grow and the rapid progress in DL for machine vision continues, we hope this will provide a first benchmark in mammogram classification on an public yet hidden test set. 

%\medskip
%\noindent\textbf{Acknowledgement:} This research was supported by the European Research Council (Starting Grant 677697,  project BUNGEE-TOOLS), by the EPSRC (CDT in Medical Imaging, EP/L016478/1), and by NVIDIA (through the donation of a Titan X GPU). 

\bibliographystyle{splncs}
\bibliography{library}

\end{document}